\documentclass{frontiersSCNS} 

\usepackage{url,hyperref,lineno,microtype}
\usepackage[onehalfspacing]{setspace}
\usepackage{times}
\usepackage{epsfig}
\usepackage{graphicx}
\usepackage{amsmath}
\usepackage{amssymb}
\usepackage{balance}

\def\keyFont{\fontsize{8}{11}\helveticabold }
\def\firstAuthorLast{Wang} 
\def\Authors{Guanghui Wang}
\def\Address{1. Department of Electrical Engineering and Computer Science, 
University of Kansas,
1520 West 15th Street
Lawrence, KS  66045\\
2. National Laboratory of Pattern Recognition, Institute of Automation, Chinese Academy of Sciences, Beijing, China 100190}
\def\corrAuthor{Guanghui Wang}
\def\corrAddress{EECS, University of Kansas}
\def\corrEmail{ghwangca@gmail.com}

\begin{document}
\onecolumn
\firstpage{1}

\title[Robust Structure and Motion Factorization]{Robust Structure and Motion Factorization of Nonrigid Objects} 

\author[\firstAuthorLast ]{\Authors} 
\address{} 
\correspondance{} 

\extraAuth{}

\maketitle


\begin{abstract}

\section{}
Structure from motion is an import theme in computer vision. Although great progress has been made both in theory and applications, most of the algorithms only work for static scenes and rigid objects. In recent years, structure and motion recovery of nonrigid objects and dynamic scenes has received a lot of attention. In this paper, the state-of-the-art techniques for structure and motion factorization of nonrigid objects are reviewed and discussed. First, an introduction of the structure from motion problem is presented, followed by a general formulation of nonrigid structure from motion. Then, an augmented affined factorization framework, by using homogeneous representation, is presented to solve the registration issue in the presence of outlying and missing data. Third, based on the observation that the reprojection residuals of outliers are significantly larger than those of inliers, a robust factorization strategy with outlier rejection is proposed by means of the reprojection residuals, followed by some comparative experimental evaluations. Finally, some future research topics in nonrigid structure from motion are discussed.

\tiny
 \keyFont{ \section{Keywords:} Structure from motion, nonrigid object, robust algorithm, matrix factorization, outlier rejection.} 
\end{abstract}

\section{Introduction}

%
Structure from motion (SfM) refers to the process of extracting three dimensional structure of the scene as well as camera motions by analyzing an image sequence. SfM is an important theme in computer vision and great progress has been made both in theory and in practice during the last three decades. Successful applications include robot navigation, augmented reality, industrial inspection, medical image analysis, digital entertainment, and many more.

The classical method for 3D reconstruction is stereo vision using
two or three images \citep{Hartley04}, where the 3D structure is
calculated via triangulations from the correspondences
between these images. For a sequence of many images, the
typical approach is the structure and motion factorization
algorithm, which was first proposed by Tomasi and Kanade
\citep{Tomasi92}. The factorization method is based on a bilinear
formulation that decomposes image measurements directly into the
structure and motion components. By assuming the
tracking matrix of an image sequence is available, the algorithm deals
uniformly with the data from all images; thus, its solution is more stable and
accurate than the stereo vision method
\citep{Ozden10}\citep{quan96}\citep{resch2015scalable}\citep{Triggs96}.

The main idea of the factorization algorithm is to decompose the
tracking matrix into the motion and structure components
simultaneously by Singular Value Decomposition (SVD) with low-rank
approximation. Most of the studies on the problem assume an affine camera model due to its linearity
\citep{Hartley08}. Christy and
Horaud \citep{Christy96} extended the method to a perspective camera
model by incrementally performing the affine factorization of a
scaled tracking matrix. Triggs \citep{Triggs96} proposed a
full projective factorization algorithm with projective depths recovered from
epipolar geometry. The method was further studied and
different iterative schemes were proposed to recover the projective
depths by minimizing image reprojection errors \citep{Oliensis07}\citep{Wang09csvt}.
Oliensis and Hartley \citep{Oliensis07} provided a complete theoretical
convergence analysis for the iterative extensions.

The factorization algorithm was extended to nonrigid SfM by assuming that the 3D
shape of a nonrigid object can be modeled as a weighted linear combination of a
set of shape bases \citep{Bregler00}. Thus, the shape bases and camera motions are
factorized simultaneously for all time instants under a rank-$3k$
constraint of the tracking matrix. The method has been extensively
investigated and developed in \citep{Brand01}
\citep{Torresani08}. Recently, Rabaud
and Belongie \citep{Rabaud08} relaxed the Bregler's assumption
and proposed a
manifold-learning framework to solve the problem. Yan and Pollefeys
\citep{Yan08} proposed a factorization approach to recover the
structure of articulated objects. Akhter
\emph{et al.} \citep{Akhter11} proposed a dual approach to describe
the nonrigid structure in trajectory space by a linear combination
of basis trajectories. Gotardo and Martinez \citep{Gotardo11} proposed to use kernels to model nonlinear deformation. More recent study can be found in \citep{agudo2014good} and \citep{newcombe2015dynamicfusion}

Most factorization methods assume that all features are tracked across the sequence.
In the presence of missing data, SVD factorization cannot be used directly, researches proposed to solve the motion and shape matrices alternatively, such as the alternative factorization \citep{KeK05}, power factorization \citep{Hartley03}, and factor analysis \citep{Gruber04}. In practice, outlying data are inevitable during the process of feature tracking, as a consequence, performance of the algorithm will degrade. The most popular strategy to handle outliers in computer vision field is RANSAC, Least Median of Squares \citep{Hartley04}, and other similar hypothesise-and-test frameworks \citep{Scaramuzza11}. However, these methods are usually designed for two or three views and they are computational expensive.

Aguitar and Moura \citep{Aguiar03} proposed a scalar-weighted SVD
algorithm that minimizes the weighted square errors. 
Gruber and Weiss \citep{Gruber04} formulated the problem
as a factor analysis and derived an Expectation Maximization (EM)
algorithm to enhance the robustness
to missing data and uncertainties. Zelnik-Manor \emph{et al.}
\citep{Zelnik-Manor06} defined a new type of motion consistency based
on temporal consistency, and applied it to multi-body factorization
with directional uncertainty. Zaharescu and Horaud
\citep{Zaharescu09} introduced a Gaussian mixture model and
incorporate it with the EM algorithm. Huynh \emph{et al.} \citep{HuynhHH03} proposed an iterative approach to correct the outliers with 'pseudo' observations. Ke and kanade \citep{KeK05} proposed a robust algorithm to handle outliers by minimizing a $L1$ norm of the reprojection errors. Eriksson and Hengel \citep{ErikssonH10} introduced the $L1$ norm to the Wiberg algorithm to handle missing data and outliers. 

Okatani \emph{et al.} \citep{OkataniYD11} proposed to incorporate a damping factor into the Wiberg method to solve the problem. Yu \emph{et al.} \citep{YuCS11} presented a Quadratic Program formulation for robust multi-model fitting of geometric structures. Wang \emph{et al.} \citep{WangCS12} proposed an adaptive kernel-scale weighted hypotheses to segment multiple-structure data even in the presence of a large number of outliers. Paladini \emph{et al.} \citep{PaladiniBXASD12} proposed an alternating bilinear approach to SfM by introducing a globally optimal projection step of the motion matrices onto the manifold of metric constraints. Wang \emph{et al.} \citep{Wang12csvt} proposed a spatial-and-temporal-weighted factorization approach to handle significant noise in the measurement. The authors further proposed a rank-4 factorization algorithm to handle missing and outlying data \citep{Wang13WACV}.

Most of the above robust algorithms are initially designed for SfM of rigid objects, few studies have been carried out for nonrigid case. In our most recent study \citep{Wang13CRV}, a robust nonrigid factorization approach is reported. The outlying data are detected from a new viewpoint via image reprojection residuals by exploring the fact that the reprojection residuals are largely proportional to the measurement errors. In this paper, the state-of-the-art techniques for structure from motion of nonrigid objects are reviewed, with some most recent development and results in this field.

The remaining part of this paper is organized as follows. Some background of structure and motion recovery of rigid objects is outlined in Section \ref{sec:background}. Section \ref{sec:nonrigid} presents the formulation and development of nonrigid SfM. An augmented factorization framework for nonrigid SfM and a robust factorization strategy are introduced in Section \ref{sec:method}, followed by some experimental results on both synthetic and real data in Section \ref{sec:results}. Finally, the paper is concluded and discussed in Section \ref{sec:con}.

\section{Structure and Motion Factorization of Rigid Objects}
\label{sec:background}

In this section, a brief review to camera projection models and rigid structure and motion
factorization are presented.

Under perspective projection, a 3D point $\mathbf{X}_j=[x_{j},
y_{j}, z_{j}]^{T}$ is projected onto an image point
$\mathbf{x}_{ij}=[u_{ij}, v_{ij}]^{T}$ in frame $i$ according to
the imaging equation
\begin{equation}\label{eq:perspective_projection}
    \lambda_{ij}\mathbf{\tilde{x}}_{ij}=\mathbf{P}_i\mathbf{\tilde{X}}_j =
    \mathbf{K}_i[\mathbf{R}_i | \mathbf{t}_i]\mathbf{\tilde{X}}_j
\end{equation}
where $\lambda_{ij}$ is a non-zero scale factor;
$\mathbf{\tilde{x}}_{ij}$ and $\mathbf{\tilde{X}}_{j}$ are the
homogeneous form of $\mathbf{{x}}_{ij}$ and $\mathbf{{X}}_{j}$,
respectively; $\mathbf{P}_i$ is a $4\times 3$ projection matrix of the $i$-th
frame; $\mathbf{R}_i$ and $\mathbf{t}_i$ are the corresponding
rotation matrix and translation vector of the camera with respect to
the world system; $\mathbf{K}_i$ is the camera calibration matrix.
When the object is far away from the camera with
relatively small depth variation, one may safely assume a simplified
affine camera model as below to approximate the perspective projection.
\begin{equation}\label{eq:affine_projection1}
    \mathbf{{x}}_{ij}=\mathbf{A}_i\mathbf{{X}}_j +
    \mathbf{{c}}_i
\end{equation}
where the matrix $\mathbf{A}_i$ is a $2\times3$ affine projection
matrix; $\mathbf{{c}}_i$ is a two-dimensional translation term of the frame. Under the affine projection, the mapping from space to the
image becomes linear as the unknown depth scalar $\lambda_{ij}$ in
(\ref{eq:perspective_projection}) is eliminated in
(\ref{eq:affine_projection1}). Consequently, the projection of all image points in the $i$-th frame can be denoted as
\begin{equation}\label{eq:ortho_i}
[\mathbf{{x}}_{i1}, \mathbf{{x}}_{i2}, \cdots, \mathbf{{x}}_{in}]
=\mathbf{A}_{i}[\mathbf{{X}}_1, \mathbf{{X}}_2, \cdots,
\mathbf{{X}}_n] + \mathbf{C}_i
\end{equation}
where $\mathbf{C}_i=[\mathbf{{c}}_i, \mathbf{{c}}_i, \cdots, \mathbf{{c}}_i]$ is the translation matrix of the frame $i$. Therefore, the imaging process of an image sequence can be formulated by stacking equation (\ref{eq:ortho_i}) frame by frame.
\begin{equation}\label{eq:affine_rigid_fact1}
\underbrace{\left[ {{\begin{smallmatrix}
 {{\mathbf{  {x}}}_{11}} & \cdots & {{\mathbf{  {x}}}_{1n} } \\
 \vdots & \ddots & \vdots \\
 {{\mathbf{  {x}}}_{m1} } & \cdots & {{\mathbf{  {x}}}_{mn} } \\
\end{smallmatrix} }} \right]}_{\mathbf{W}_{2m\times n}} = \underbrace{\left[{\begin{smallmatrix}
 \mathbf{A}_{1} \\
 \vdots \\
 \mathbf{A}_{m} \\
\end{smallmatrix} }\right]}_{\mathbf{M}_{2m\times 3}} \underbrace{\left[{\begin{smallmatrix}
\mathbf{ {X}}_1, & \cdots, & \mathbf{ {X}}_n
\end{smallmatrix} }\right]}_{\mathbf{{S}}_{3\times n}} + \underbrace{\left[{\begin{smallmatrix}
 \mathbf{C}_{1} \\
 \vdots \\
 \mathbf{C}_{m} \\
\end{smallmatrix} }\right]}_{ \mathbf{C}_{2m\times n}}
\end{equation}
where $m$ is the frame number and $n$ is the number of features.
It is easy to verify that $\mathbf{{c}}_i$ in (\ref{eq:affine_projection1}) is the image of the centroid of all space points. Thus, if all imaged points in each image are
registered to the centroid and relative image coordinates with respect to the centroid are
employed, the translation term vanishes, i.e. $\mathbf{{c}}_i=\mathbf{0}$. 
Consequently, the imaging process (\ref{eq:affine_rigid_fact1}) is written concisely as
\begin{equation}\label{eq:sim_affine}
\mathbf{W}_{2m\times n}=\mathbf{M}_{2m\times 3}
\mathbf{{S}}_{3\times n}
\end{equation}
where $\mathbf{W}$ is called the tracking matrix or measurement matrix, which is composed by all tracked features. Structure from motion is a reverse problem to the imaging process. Suppose the
tracking matrix $\mathbf{W}$ is available, our purpose is to recover the motion matrix
$\mathbf{M}$ and the shape matrix $\mathbf{{S}}$.

It is obvious from (\ref{eq:sim_affine}) that the tracking matrix is highly
rank-deficient, and the rank of $\mathbf{W}$ is at
most 3 if the translation term $\mathbf{C}$ is removed. In practice, the rank of a real tracking matrix is definitely greater
than 3 due to image noise and affine approximation error. Thus, one need to find a rank-3 approximation of the
tracking matrix. A common practice is to perform SVD decomposition
on matrix $\mathbf{W}$ and truncate it to rank 3, then the motion
matrix $\mathbf{M}$ and the shape matrix $\mathbf{{S}}$ can be
easily decomposed from the tracking matrix. Nevertheless, this
decomposition is not unique since it is only defined up to a
nonsingular linear transformation matrix $\mathbf{H}\in
\mathbb{R}^{3\times 3}$ as
\begin{equation}\label{eq:up}
    \mathbf{W}=({\mathbf{{M}H}})({\mathbf{H}}^{-1}{\mathbf{{S}}})
\end{equation}

In order to upgrade the solution from perspective space to
the Euclidean space, the metric constraint on the motion matrix is
usually adopted to recover the transformation matrix $\mathbf{H}$
\citep{quan96}\citep{WangIJCV10}. Then, the Euclidean structure is recovered from ${\mathbf{H}}^{-1}{\mathbf{{S}}}$ and the camera motion parameters of each frame are decomposed from $\mathbf{{M}H}$.

\section{Structure and Motion Factorization of Nonrigid Objects}
\label{sec:nonrigid}

We assumed rigid objects and static scenes in last section. While in
real world, many objects do not have fixed structure, such as human
faces with different expressions, torsos, and animals bodies, etc.
In this section, the factorization algorithm is extended to handle
nonrigid and deformable objects. 

\subsection {Bregler's deformation model}

\begin{figure}[t]
  \centering
  \includegraphics[width=0.7\textwidth]{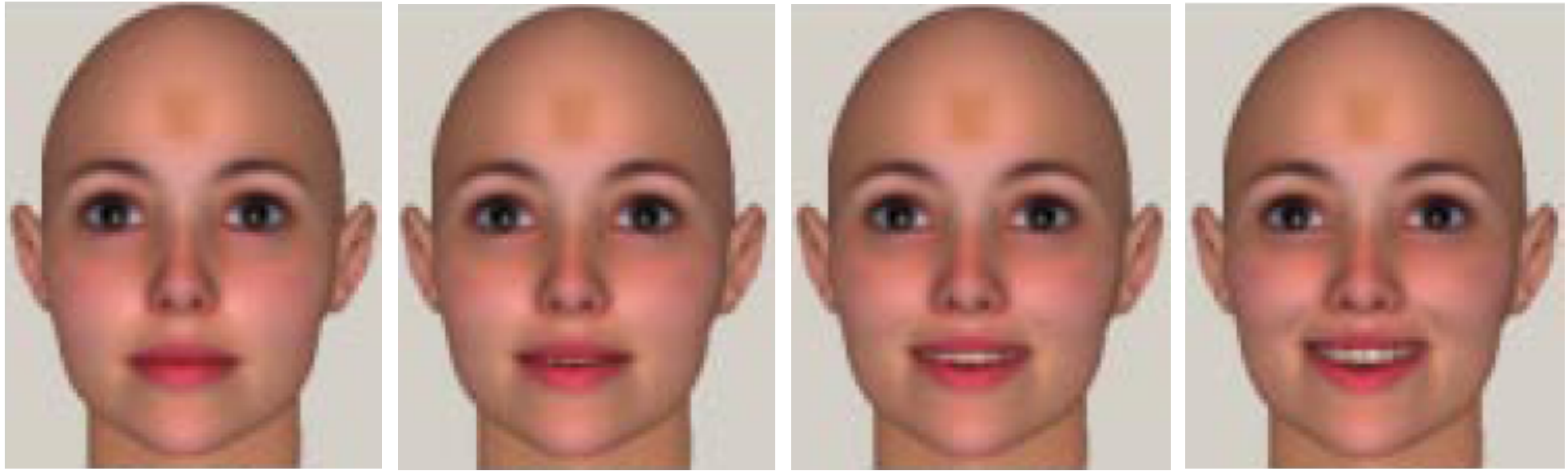}\\
  \caption{Four female face models carrying expressions from neutral to smiling.
  We may take any two models as shape bases,
  then the other models can be derived as weighted linear combinations of the
  two bases. Courtesy of Jing Xiao.}\label{fig:nonface}
\end{figure}

For nonrigid objects, if their surfaces deform randomly at any
time instance, there is currently no suitable method to recover its
structure from images. A well adopted assumption about the deformation is proposed by Bregler
 \citep{Bregler00}, where the 3D structure of nonrigid
object is approximated by a weighted combination of a set of rigid shape
bases. Fig. \ref{fig:nonface} shows a very simple example of face
models from neutral to smiling with only mouth movements. The
deformation structure can be approximated by only two shape bases. If
more face expressions, such as joy, sadness, surprise, fear, etc.,
are involved, then more \index{shape bases} shape bases are needed
to model the structure.

Suppose the \index{deformation structure} deformation structure
$\mathbf{{S}}_i \in \mathbb{R} ^{3\times n}$ is expressed as a
weighted combination of $k$ principal modes of deformation
$\mathbf{B}_l \in \mathbb{R} ^{3\times n}, l=1, \cdots, k$. We
formulate the model as
\begin{equation}\label{eq:bregler}
\mathbf{{S}}_i = \sum_{l = 1}^k {\omega_{il}} \mathbf{B}_l
\end{equation}
where $\omega_{il} \in \mathbb{R} $ is the deformation weight for
base $l$ at frame $i$. A perfect rigid object corresponds to the
situation of $k = 1$ and $\omega_{il}=1$. 

\subsection {Nonrigid factorization under affine models}\index{nonrigid factorization}

Under the assumption (\ref{eq:bregler}), the imaging process of one image can be modeled as
\begin{eqnarray}\label{eq:nonrigid_ortho_i}
\mathbf{W}_i &=& [\mathbf{x}_{i1},
\cdots, \mathbf{x}_{in}] =\mathbf{A}_{i}\mathbf{S}_i+[\mathbf{c}_{i},
\cdots, \mathbf{c}_{i}]\nonumber\\
&=& \left[ {\omega_{i1} \mathbf{A}_{i}}, \cdots, {\omega_{ik}
\mathbf{A}_{i}} \right] \left[\begin{smallmatrix} \mathbf{B}_1 \\ \vdots \\
\mathbf{B}_k \\
\end{smallmatrix}\right]+ {[\mathbf{c}_{i},
\cdots, \mathbf{c}_{i}]}\nonumber
\end{eqnarray}

It is easy to verify that if all image points in each frame are
registered to the centroid and relative image coordinates are
employed, the translation term vanishes, i.e., $\mathbf{{c}}_i=\mathbf{0}$.
Consequently, the nonrigid factorization under affine camera model is
expressed as
\begin{equation}\label{eq:nonrigid_fact}
    \underbrace{\left[ {{\begin{smallmatrix}
 {{\mathbf{ x}}_{11}} & \cdots & {{\mathbf{ x}}_{1n} } \\
 \vdots & \ddots & \vdots \\
 {{\mathbf{ x}}_{m1} } & \cdots & {{\mathbf{ x}}_{mn} } \\
\end{smallmatrix} }} \right]}_{\mathbf{W}_{2m\times n}} = \underbrace{\left[
{{\begin{smallmatrix}
 {\omega_{11} \mathbf{A}_1} & \cdots &
{\omega_{1k} \mathbf{A}_1} \\
 \vdots & \ddots & \vdots \\
 {\omega_{m1} \mathbf{A}_m} & \cdots &
{\omega_{mk} \mathbf{A}_m} \\
\end{smallmatrix} }} \right]}_{\mathbf{M}_{2m\times 3k}} \underbrace{\left[\begin{smallmatrix} \mathbf{B}_1 \\ \vdots \\
\mathbf{B}_k \\
\end{smallmatrix}\right]}_{\mathbf{S}_{3k\times n}}
\end{equation}

Structure from motion is a reverse problem. Suppose the
tracking matrix $\mathbf{W}$ is available, our purpose is to recover the camera motion parameters in $\mathbf{M}$ and the 3D structure from the shape matrix $\mathbf{S}_i$. It is obvious from (\ref{eq:nonrigid_fact}) that the rank of the tracking matrix $\mathbf{W}$ is at most $3k$. 

Following the idea of rigid factorization, we perform SVD
decomposition on the nonrigid tracking matrix and impose the
rank-$3k$ constraint, $\mathbf{W}$ can be factorized into a
$2m\times 3k$ matrix $\mathbf{\hat{M}}$ and a $3k\times n$ matrix
$\mathbf{\hat{S}}$. However, the decomposition is not unique as
any nonsingular linear transformation matrix $\mathbf{H} \in
\mathbb{R}^{3k\times 3k}$ can be inserted into the factorization
which leads to an alternative factorization
$\mathbf{W}=(\mathbf{\hat{M}}\mathbf{H})(\mathbf{H}^{-1}\mathbf{\bar{\hat{S}}})$.
If we have a transformation matrix ${\rm {\bf H}}$ that can resolve
the affine ambiguity and upgrade the solution to Euclidean space,
the shape bases are then easily recovered from ${\rm {\bf S}} = {\rm
{\bf H}}^{ - 1}{\rm {\bf \hat {S}}}$, while the rotation matrix
 and the weighting coefficient $\omega_{ij} $
can be decomposed from ${\rm {\bf M}} = {\rm {\bf \hat {M}H}}$ by
Procrustes analysis
\citep{Brand01}\citep{Bregler00}\citep{Torresani01}. 

Similar to the
rigid situation, the \index{upgrading matrix} upgrading matrix is
usually recovered by application of metric constraint to the motion
matrix \citep{WangIJCV10}. However, only the rotation
constraints may be insufficient when the object deforms at varying
speed, since most of the constraints are redundant. Xiao \emph{et
al.} \citep{Xiao05} proposed a \index{basis constraint} basis
constraint to solve this ambiguity. The main idea is based on the assumption that there exists $k$
frames in the sequence which include independent shapes that can be
treated as a set of bases.

\section{Material and Methods}\label{sec:method}
In nonrigid structure from motion, the input is an image sequence of a nonrigid object with the point features being tracked across the sequence. However, due to occlusion and lack of proper constraints, the tracking data is usually corrupted by outliers and missing points. This section will introduce a robust scheme to handle imperfect data \citep{Wang13CRV}.

\subsection{Augmented affine factorization without registration}
\label{sec:robust}
One critical condition for the affine factorization equation (\ref{eq:nonrigid_fact}) is that all image measurements are registered to the corresponding centroid of each frame. When the tracking matrix contains outliers and/or missing data, it is impossible to reliably retrieve the centroid. As will be shown in the experiments, the miscalculation of the centroid will cause a significant error to the final solution. Previous studies were either ignoring this problem or hallucinating the missing points with pseudo observations, which may lead to a biased estimation. In this section, a rank-($3k+1$) augmented factorization algorithm is proposed to solve this issue.

Let us formulate the affine imaging process (\ref{eq:affine_projection1}) in the following form
\begin{equation}\label{eq:affine_projection4}
    \mathbf{{x}}_{ij}=\left[ \mathbf{A}_i | \mathbf{{c}}_i \right]\mathbf{{\tilde{X}}}_j
\end{equation}
where $\mathbf{{\tilde{X}}}_j=[\mathbf{X}_j^T, t_j]^T$ is a 4-dimensional homogeneous expression of $\mathbf{{X}}_j$. Let $\mathbf{\tilde{S}}_i=\left[{{\begin{smallmatrix}
{{\mathbf{S}}_{i}}\\\mathbf{t}_i\end{smallmatrix} }} \right]$ be the homogeneous form of the deformable structure, then the imaging process of frame $i$ can be written as
\begin{eqnarray}\label{eq:nonrigid_frame_i}
\mathbf{W}_i &=& [\mathbf{x}_{i1},
\cdots, \mathbf{x}_{in}] =\left[ \mathbf{A}_i | \mathbf{{c}}_i \right]\left[{{\begin{smallmatrix}
{\sum\nolimits_{l = 1}^k {\omega_{il}}
\mathbf{B}_l}\\\mathbf{t}_i\end{smallmatrix} }} \right] \nonumber\\
&=& \left[ {\omega_{i1} \mathbf{A}_{i}}, \cdots, {\omega_{ik}
\mathbf{A}_{i}}, \mathbf{{c}}_i \right] \left[\begin{smallmatrix} \mathbf{B}_1 \\ \vdots \\
\mathbf{B}_k \\ \mathbf{t}_i
\end{smallmatrix}\right]\nonumber
\end{eqnarray}

Thus, the structure and motion factorization for the entire sequence is formulated as follows.
\begin{equation}\label{eq:affine_rigid_fact4}
\underbrace{\left[ {{\begin{smallmatrix}
 {{\mathbf{  {x}}}_{11}} & \cdots & {{\mathbf{  {x}}}_{1n} } \\
 \vdots & \ddots & \vdots \\
 {{\mathbf{  {x}}}_{m1} } & \cdots & {{\mathbf{  {x}}}_{mn} } \\
\end{smallmatrix} }} \right]}_{\mathbf{W}_{2m\times n}} = \underbrace{\left[{\begin{smallmatrix}
 {\omega_{11} \mathbf{A}_{1}} & \cdots & {\omega_{1k} \mathbf{A}_{1}} & \mathbf{c}_{1} \\
 \vdots & \ddots & \vdots & \vdots\\
 {\omega_{m1} \mathbf{A}_{m}} & \cdots & {\omega_{mk} \mathbf{A}_{m}} & \mathbf{c}_{m} \\
\end{smallmatrix} }\right]}_{\mathbf{M}_{2m\times (3k+1)}} \underbrace{\left[\begin{smallmatrix} \mathbf{B}_1 \\ \vdots \\
\mathbf{B}_k \\ \mathbf{t}_i^T
\end{smallmatrix}\right]}_{\mathbf{{S}}_{(3k+1)\times n}}
\end{equation}

Obviously, the rank of the tracking matrix becomes $3k+1$ in this case. Given the tracking matrix, the factorization can be easily obtained via SVD decomposition and imposing rank-($3k+1$) constraint. The expression (\ref{eq:affine_rigid_fact4}) does not require any image registration thus can directly work with outlying and missing data.

Both factorization algorithms (\ref{eq:nonrigid_fact}) and (\ref{eq:affine_rigid_fact4}) can be equivalently denoted as the following minimization scheme.
\begin{equation}\label{eq:fact_Fnorm}
f(\mathbf{M},\mathbf{S})=arg\min_{\mathbf{M},\mathbf{S}}
\|\mathbf{W}-\mathbf{M}\mathbf{S}\|_F^2
\end{equation}

By enforcing different rank constraints, the Frobenius norm of (\ref{eq:fact_Fnorm}) corresponding to the algorithms (\ref{eq:nonrigid_fact}) and (\ref{eq:affine_rigid_fact4}) would be
\begin{equation}\label{eq:error_3kand+1}
E_{3k}=\sum_{i=3k+1}^N \sigma_i^2,\;\; E_{3k+1}=\sum_{i=3k+2}^N \sigma_i^2
\end{equation}
where $\sigma_i, i=1,\cdots, N$ are singular values of the tracking matrix in descending order, and $N=\min(2m, n)$. Clearly, the error difference by the two algorithm is $\sigma_{3k+1}^2$. For noise free data, if all image points are registered to the centroid, then, $\sigma_{i}=0, \forall i>3k$, the equations (\ref{eq:nonrigid_fact}) and (\ref{eq:affine_rigid_fact4}) are actually equivalent. However, in the presence of outlying and missing data, the image centroid cannot be accurately recovered, the rank-$3k$ algorithm (\ref{eq:nonrigid_fact}) will yield a big error since $\sigma_{3k+1}$ does not approach zero in this situation.


\subsection{Outlier detection and robust factorization}\label{sec:outliers}

Based on the foregoing proposed factorization algorithm, A fast and practical scheme for outlier detection is discussed in this section.


%
%

The best fit model of the factorization algorithm is obtained by minimizing the sum of squared residuals between the observed data and the fitted values provided by the model. Extensive empirical studies show that the algorithm usually yields reasonable solutions even in the presence of certain amount of outliers, and the reprojection residuals of the outlying data are usually significantly larger than those associated with inliers.
\begin{figure}[t]
  \centering
  (a)\;\; \includegraphics[width=0.5\linewidth]{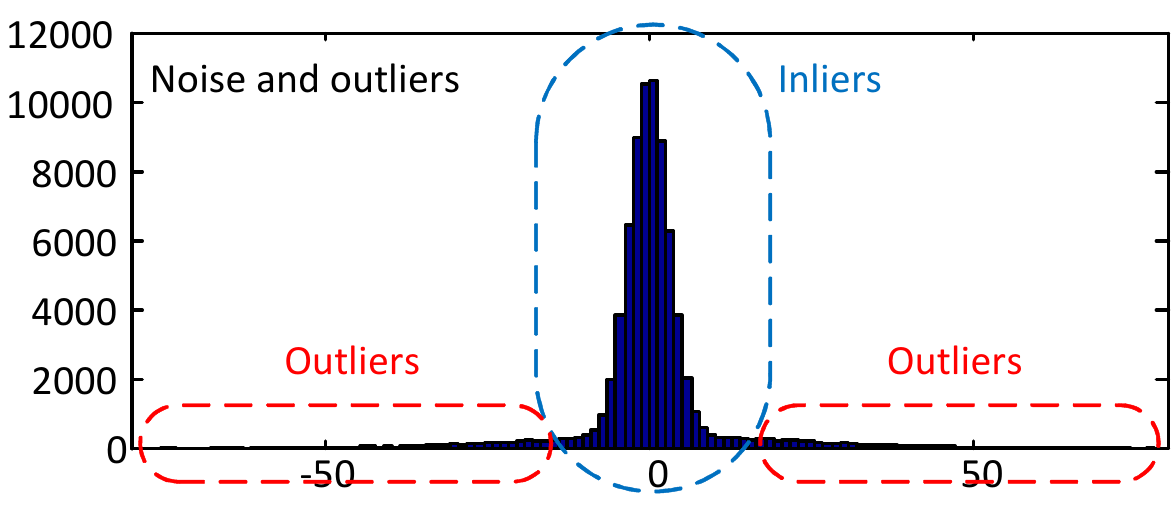}\\
  (b)\;\; \includegraphics[width=0.5\linewidth]{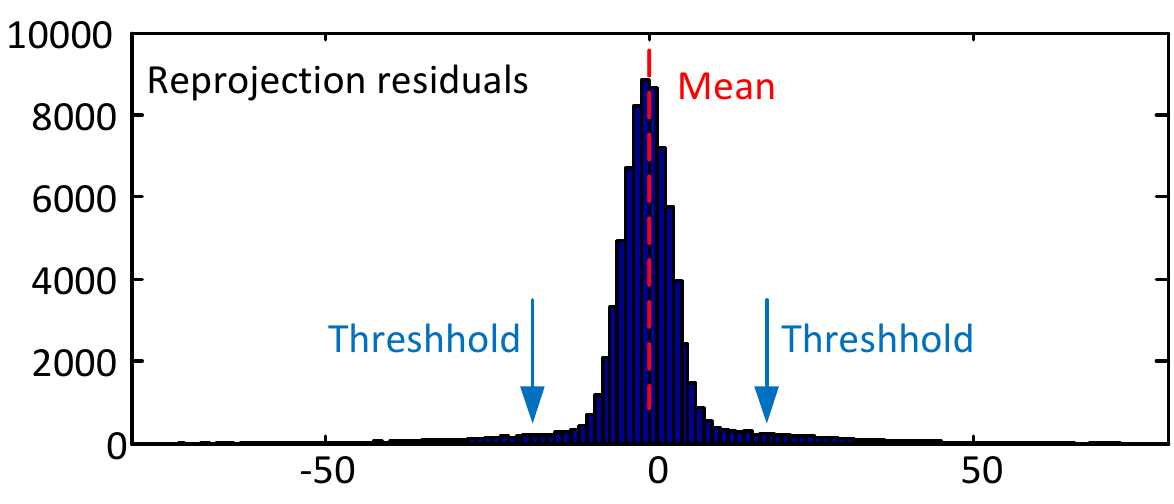}\\
  \caption{(a) Histogram distribution of the added noise and outliers; (b) histogram distribution of the reprojection residuals.}
\label{fig:s1}
\end{figure}

Suppose $\mathbf{\hat{M}}$ and $\mathbf{\hat{S}}$ are a set of initial solution of the motion and structure matrices, the reprojection residuals can be computed by
reprojecting the solution back onto all images. Let us define a residual matrix as follows.
\begin{equation}\label{eq:residual_matrix}
\mathbf{E}=\mathbf{W}-\mathbf{\hat{M}}\mathbf{\hat{S}}=
\left[\begin{smallmatrix}
 \mathbf{e}_{11} & \cdots & \mathbf{e}_{1n} \\
 \vdots & \ddots & \vdots \\
 \mathbf{e}_{m1} & \cdots & \mathbf{e}_{mn} \\
\end{smallmatrix} \right]_{2m\times n}
\end{equation}
where
\begin{equation}\label{eq:point_residual}
\mathbf{e}_{ij}=\mathbf{x}_{ij}-
\mathbf{\hat{M}}_{i}\mathbf{\hat{s}}_j=\left[\begin{smallmatrix} \Delta u_{ij}\\ \Delta v_{ij} \end{smallmatrix}\right]
\end{equation}
is the residual of point $(i,j)$ in both image directions. The reprojection error of a point is defined by the Euclidean norm of the residual at that point as $\|\mathbf{e}_{ij}\|$. 

Assuming Gaussian image noise, it is easy to prove that the reprojection residuals also follow  Gaussian distribution \citep{Wang13CRV}. Fig.\ref{fig:s1} shows an example from the synthetic data in Section \ref{sec:experiment1}. 3 units Gaussian noise and 10\% outliers were added to the synthetic images, and the residual matrix was calculated from (\ref{eq:residual_matrix}). As shown in Fig.\ref{fig:s1}, the residuals are obviously follow Gaussian distribution.
Thus, the points with large residuals will be classified as outliers.

Inspired by this observation, a simple outlier detection and robust factorization scheme is proposed below.

\noindent\rule{0.47\textwidth}{.1pt}\\
\textbf{Robust Nonrigid Factorization Algorithm} \\[-6pt]
\rule{0.47\textwidth}{.1pt}

\footnotesize
\textsf{\textbf{Input:} Tracking matrix of the sequence}


\textsf{1. Perform rank-($3k+1$) affine factorization on the tracking matrix to
  obtain an initial solution of $\mathbf{\hat{M}}$ and $\mathbf{\hat{S}}$.}

\textsf{2. Estimate the reprojection residuals (\ref{eq:residual_matrix}) and determine an outlier threshold.}

\textsf{3. Eliminate the outliers and reestimate the reprojection residuals using alternative factorization \citep{Wang13CRV}.}

\textsf{4. Redetermine the outlier threshold and eliminate the outliers.}

\textsf{5. Estimate the weight matrix $\mathbf{\Sigma}$ and perform a weighted factorization using the inliers to obtain a set of refined solution.}

\textsf{6. Recover the upgrading matrix $\mathbf{H}$ and upgrade the solution to the Euclidean space: $\mathbf{M}=\mathbf{\hat{M}}\mathbf{H}$,
$\mathbf{S}=\mathbf{H}^{-1}\mathbf{\hat{S}}$.}

\textsf{7. Recover the Euclidean structure $\mathbf{S}_i$ and motion parameters corresponding to each frame.}

\textsf{\textbf{Output:} 3D structure and camera motion parameters}
\\[-6pt]
\rule{0.47\textwidth}{.1pt}\\[-6pt]
\normalsize

Two important parameters are required in the robust algorithm: one is the outlier threshold, the other is the weight matrix. A detailed discussion on how to recover these parameters can be found in \citep{Wang13CRV}. 

\begin{figure}[t]
  \centering
  \includegraphics[width=0.9\linewidth]{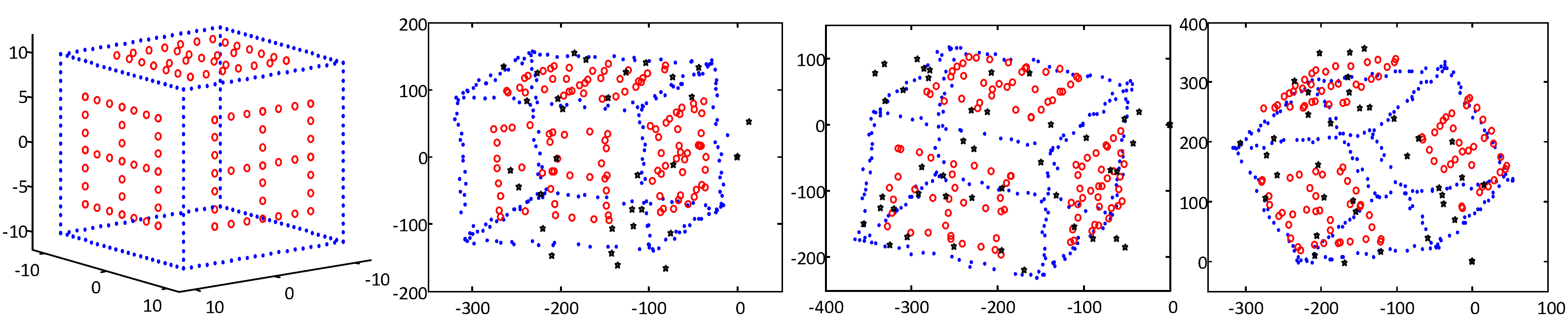}\\[-3pt]%
  (a) \hspace{0.2\linewidth}(b)\hspace{0.21\linewidth}(c)\hspace{0.2\linewidth}(d)\\
  \includegraphics[width=0.9\linewidth]{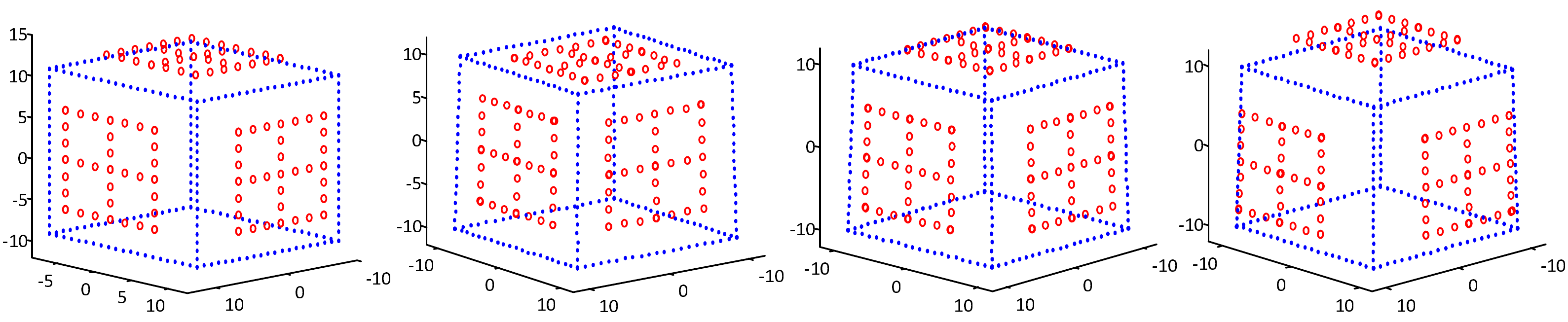}\\[-3pt]%
  (d) \hspace{0.2\linewidth}(e)\hspace{0.21\linewidth}(f)\hspace{0.2\linewidth}(g)\\
  \caption{(a) (d) Two simulated space cubes with three sets of moving points; (b) (c) (d) three synthetic images with noise and outliers (black stars); (e) (f) (g) the reconstructed 3D structures corresponding to the three images.}
\label{fig:s2}
\end{figure}
\begin{figure}[t]
  \centering
  \includegraphics[width=0.9\linewidth]{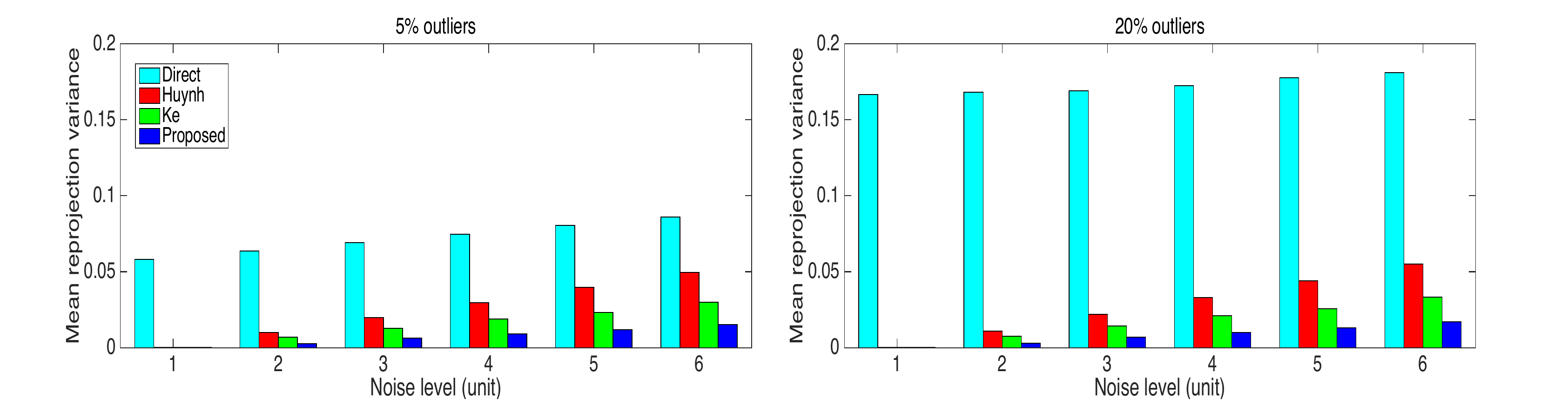}\\[-3pt]%
  (a) \hspace{0.4\linewidth}(b)\\
  \caption{The mean reprojection variance of different algorithms with respect to different noise levels and outliers. (a) 5\% outliers; (b) 20\% outliers.} \label{fig:s5}
\end{figure}

\section{Evaluations on Synthetic and Real Data}
\label{sec:results}

The section presents two examples of structure and motion recovery using the above robust factorization scheme.

\subsection{Evaluations on synthetic data}
\label{sec:experiment1}

The proposed technique was evaluated extensively on
synthetic data and compared with previous algorithms. During the
simulation, we generated a deformable space cube, which was composed of 21 evenly distributed rigid points on each side and three sets of dynamic points ($33\times 3$ points) on the
adjacent surfaces of the cube that were moving outward. There are 252 space points in total as shown in Fig.\ref{fig:s2}. Using the synthetic cube, 100 images was generated
by affine projection with randomly selected camera parameter. Each image corresponds to a different 3D structure. The image resolution is $800\times 800$ unit and Gaussian white noise is added to the synthetic images.



For the above simulated image sequence, Gaussian noise was added to each image point and the noise level was varied from 1 unit to 5 units in steps of 1. In the mean time, 10\% outliers were added to the tracking matrix. Using the contaminated data, the foregoing proposed robust algorithm was employed to recover the motion and shape matrices. Fig.\ref{fig:s2} shows three noise and outlier corrupted images and the corresponding 3D structures recovered by the proposed approach. It is evident that the deformable cube structures are correctly retrieved.

As a comparison, two popular algorithms in the literature were implemented as well, one is an outlier correction scheme proposed by Huynh \emph{et al.} \citep{HuynhHH03}, the other one is proposed by Ke and kanade \citep{KeK05} based on minimization of the $L1$ norm. The two algorithms were initially proposed for rigid SfM, here they were extended to deal with nonrigid SfM. The mean reprojection variance at different noise levels and outliers ratios is shown in Fig.\ref{fig:s5}.

The results in Fig.\ref{fig:s5} were evaluated from 100 independent tests in order to yield a statistically meaningful results, where 'Direct' stands for normal factorization algorithm without outlier rejection. In this test, the reprojection variance was estimated only using the original inlying data without outliers. Obviously, the proposed scheme outperforms other algorithms in terms of accuracy. The direct factorization algorithm yields significantly large errors due to the influence of outliers, and the error increases with the increase of the amount of outliers. The experiment also shows that all three robust algorithms are resilient to outliers, as can be seen in Fig. \ref{fig:s5}, the ratio of outliers has little influence to the reprojection variance of the three robust algorithms.


\subsection{Evaluations on real sequences}
\label{sec:experiment2}

\begin{figure}[t]
  \centering
  \includegraphics[width=0.9\linewidth]{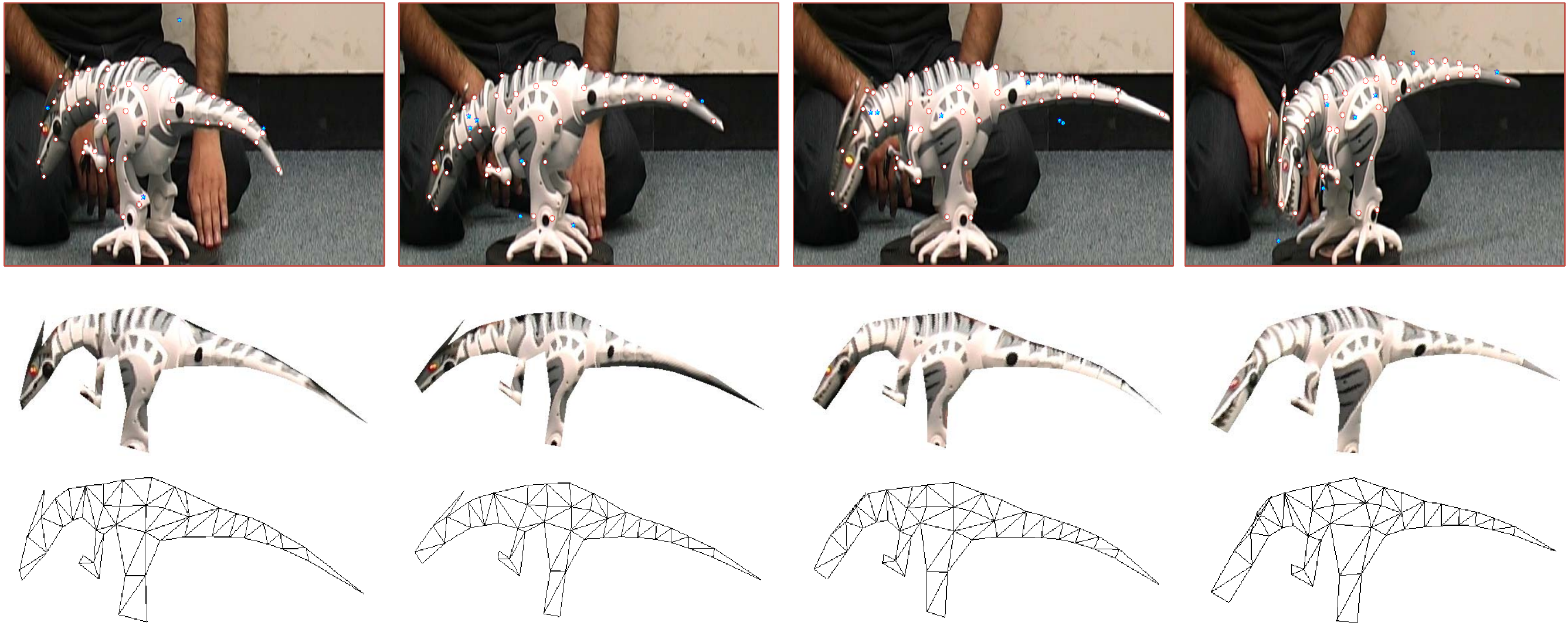}\\
  \caption{Four frames from the dinosaur sequence, the corresponding reconstructed VRML models, and  triangulated wireframes. The tracked features (red circles) and added outliers (blue stars) are superimposed to the images.} \label{fig:r1}
\end{figure}

Two experimental results are reported in this section to test the robust scheme. The first one was on a dinosaur sequence sequence \citep{Akhter11} is reported here. The sequence consists of 231 images with different movement and deformation of a dinosaur model. The image resolution is $570\times 338$ pixel and 49 features were tracked across the sequence. In order to test the robustness of the algorithm, an additional 8\% outliers were added to the tracking data as shown in Fig.\ref{fig:r1}. 

Using the proposed approach, all outliers were successfully rejected, however, a few tracked features were also eliminated due to large tracking errors. The proposed approach was employed to recover the motion and structure matrices, and the solution was then upgraded to the Euclidean space. Fig.\ref{fig:r1} shows the reconstructed structure and wireframes. The VRML model is visually realistic and the deformation at different instant is correctly recovered, although the initial tracking data are not very reliable.
\begin{figure}[t]
  \centering
  \includegraphics[width=0.85\linewidth]{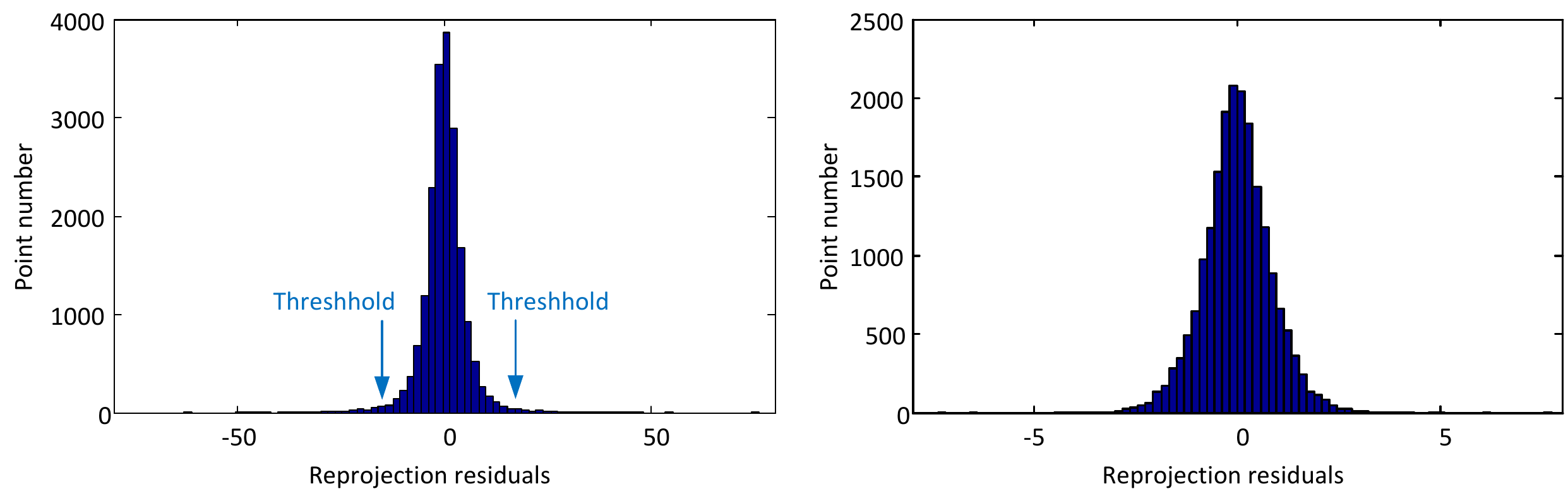}\\
  \caption{The histogram distribution of the residual matrix of the dinosaur sequence before (a) and after (b) outlier rejection.} \label{fig:r2}
\end{figure}

The histogram distribution of the reprojection residual matrix (\ref{eq:residual_matrix}) with outliers is shown in Fig.\ref{fig:r2}. The residuals are largely conform to the assumption of normal distribution. As can be seen from the histogram, the outliers are obviously distinguished from inliers, the computed threshold is shown in the figure. After rejecting outliers, the histogram distribution of the residuals produced by the final solution is also shown in Fig.\ref{fig:r2}. Clearly, the residual error is reduced significantly. The final mean reprojection error given by the proposed approach is 0.597. In comparison, the reprojection errors by the algorithms of 'Huynh' and 'Ke' are 0.926 and 0.733, respectively. The proposed scheme outperforms other approaches.

\begin{figure}[t]
  \centering
  \includegraphics[width=0.9\linewidth]{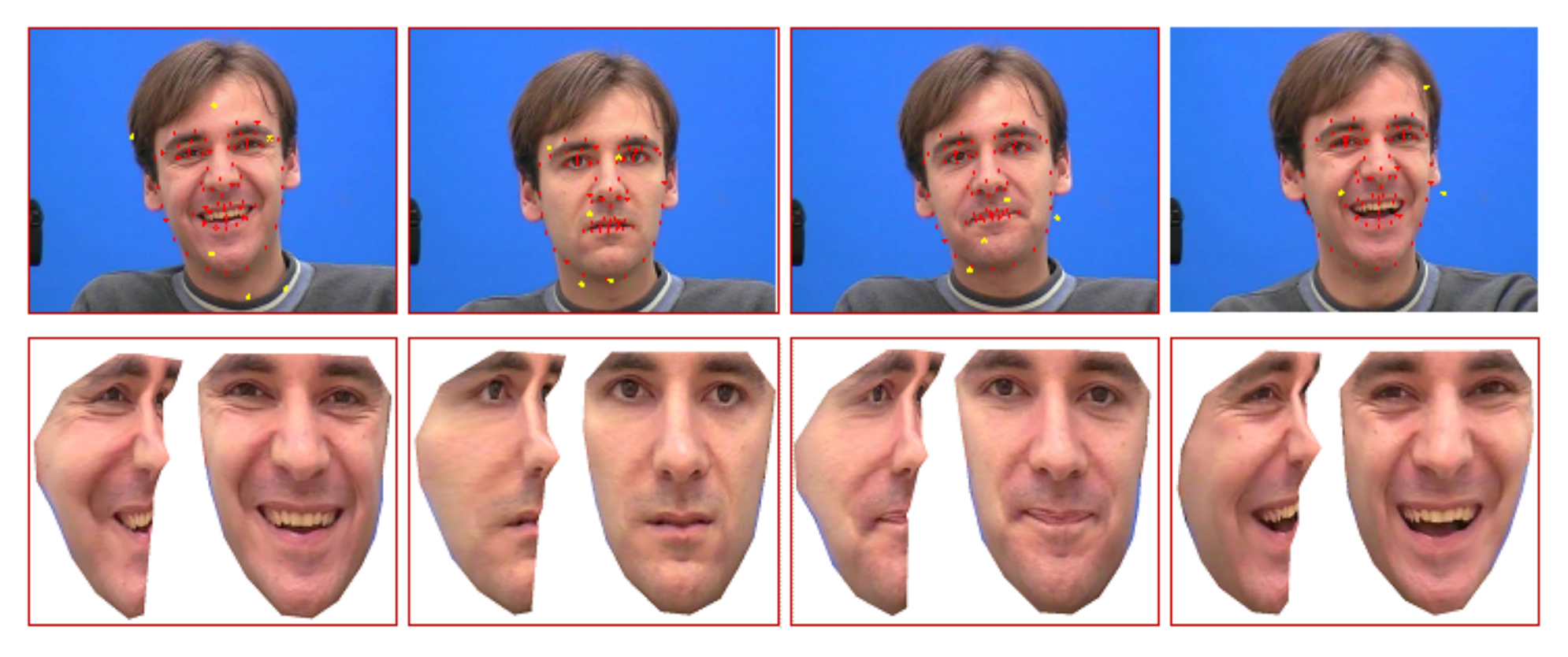}\\
  \caption{Test results of the face sequence. (top) Four frames from the sequence overlaid with the tracked features (red circles) and added outliers (blue stars); (bottom) the corresponding 3D VRML models from two different viewpoints.} \label{fig:r3}
\end{figure}

The second test was on a nonrigid face sequence,  as shown in Fig.\ref{fig:r3}, with different facial expressions. This sequence was downloaded from FGnet (http://www-prima.inrialpes.fr/FGnet/html/home.html), and 200 consecutive images were used in the experiment. The resolution of each image is $720\times 576$ and 68 feature points are automatically tracked across the sequence using the active appearance model (AAM). In order to test the robustness of the approach, 8\% outliers were added radomely to the tracking data as shown in Fig.\ref{fig:r3}. 

Fig.\ref{fig:r3} shows the reconstructed VRML models of four frames shown from front and right side. It is obvious that all outliers are removed successfully by the proposed algorithm, and different facial expressions have been correctly recovered. The reprojection errors obtained from 'Huynh', 'Ke', and the proposed algorithms are 0,697, 0.581, and 0.453, respectively. The proposed scheme again yields the lowest reprojection error in this test.

\section{Conclusion and Discussion}
\label{sec:con}

The paper has presented an overview of the state-of-the-art techniques for nonrigid structure and motion factorization. A rank-($3k+1$) factorization algorithm, which is more accurate and more widely applicable than the classic rank-$3k$ nonrigid factorization, has been presented, followed by a robust factorization scheme designed to deal with outlier-corrupted data. Comparative experiments demonstrated the effective and robustness of the proposed scheme.
Although an outlier detection strategy and a robust factorization approach have been discussed in this paper, the problem remains far from being solved. Robustness is still a well-known bottleneck in practical applications of the algorithm. On the other hand, the nonrigid factorization approach discussed in this paper only works for a single simply nonrigid object, how to recover the 3D structure of general dynamic scenes, which may be coupled with rigid, nonrigid, articulated, and moving objects, is still an open problem in computer vision. New techniques have to be developed to handle the complex and challenging situations.

\section*{Acknowledgments}

The work is partly supported by Kansas NASA EPSCoR Program and the National Natural Science Foundation of China (61273282, 61573351).



\end{document}